\documentclass[10pt,twocolumn,letterpaper]{article}

\usepackage{cvpr}              

\usepackage{graphicx}
\usepackage{amsmath}
\usepackage{amssymb}
\usepackage{booktabs}

\usepackage[pagebackref,breaklinks,colorlinks]{hyperref}

\usepackage[capitalize]{cleveref}
\crefname{section}{Sec.}{Secs.}
\Crefname{section}{Section}{Sections}
\Crefname{table}{Table}{Tables}
\crefname{table}{Tab.}{Tabs.}

\def\cvprPaperID{55} 
\def\confName{AI4CC @ CVPR}
\def\confYear{2024}

\begin{document}

\title{My Body My Choice: Human-Centric Full-Body Anonymization}

\author{Umur Aybars \c{C}ift\c{c}i\\
Binghamton University\\
{\tt\small uciftci@binghamton.edu}
\and
Ali Kemal Tanr{\i}verdi\\
Binghamton University\\
{\tt\small atanriv1@binghamton.edu}
\and
\.{I}lke Demir\\
Intel Labs\\
{\tt\small ilke.demir@intel.com}
}
\twocolumn[{\renewcommand\twocolumn[1][]{#1}%
\maketitle
\vspace{-0.5in}
\begin{center}
    \includegraphics[width=1\textwidth]{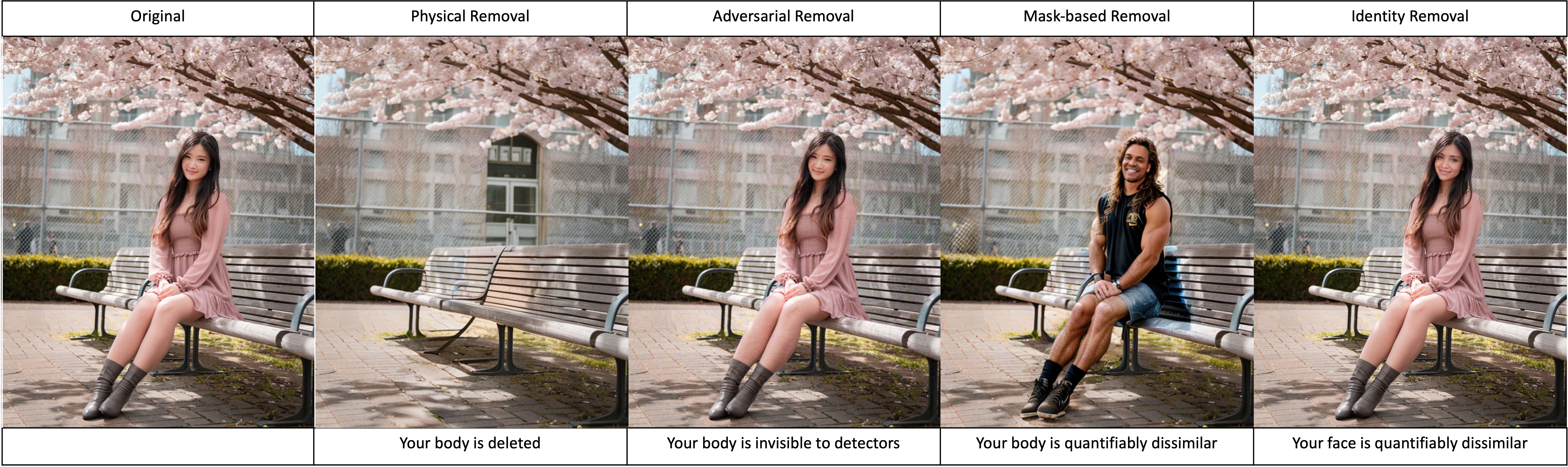}\\
    \textbf{Figure 1. }{\textbf{My Body My Choice.} Four anonymization options are provided for choosing if and how to appear in a photo.}
    \label{fig:teaser}
\end{center}%
\stepcounter{figure}
}]

\begin{abstract}
In an era of increasing privacy concerns for our online presence, we propose that the decision to appear in a piece of content should only belong to the owner of the body. Although some automatic approaches for full-body anonymization have been proposed, 
human-guided anonymization can adapt to various contexts, such as cultural norms, personal relations, esthetic concerns, and security issues. ``My Body My Choice’’ (MBMC) enables physical and adversarial anonymization by removal and swapping approaches aimed for four tasks, designed by single or multi, ControlNet or GAN modules, combining several diffusion models. We evaluate anonymization on seven datasets; compare with SOTA inpainting and anonymization methods; evaluate by image, adversarial, and generative metrics; and conduct reidentification experiments.
\end{abstract}

\section{Introduction} 
In the digital age, the proliferation of social media and the ubiquity of data collection have given rise to profound concerns about privacy and security~\cite{ziegeldorf2014privacy}. 
In physical reality, we \textit{choose} to be somewhere or not, or we \textit{choose} to be seen somewhere or not. In digital world, anyone with a photo or likeness of someone can freely publish that content to be seen by, well, everyone. We introduce ``My Body My Choice'' (MBMC), so whenever a photo of you is uploaded, you can \textit{choose} to be in it and how you want to appear in it. 

MBMC recognizes that anonymization is not a one-size-fits-all endeavor, therefore the system offers four anonymization options (Fig.1~\ref{fig:teaser}). First, the body can be physically deleted without breaking the image integrity. Second, the body can be adversarially deleted so that person detectors cannot process it automatically, while friends and family observes no difference in the image. Third, the body can be replaced by a quantifiably dissimilar body as an anonymization mask. Fourth, the identity of the body can be removed with minimal changes to the rest of the body. 
Our contributions include, a human-centric full-body anonymization paradigm enabling people to choose if and how they appear in photos, and a full-body anonymization algorithm that masks the input body with quantifiably most dissimilar body without breaking image continuity.

We extensively evaluate our approach, starting by collecting diverse images from seven different datasets, spanning various poses, activities, demographics, scenes, contexts, crowds, and environments. 
We evaluate the anonymized images in pixel, noise, generative, adversarial, visual, and structural spaces. We believe that on the rise of global privacy regulations, such systems will be the cornerstone of social media platforms.

\begin{figure*}[ht]
  \centering
  \includegraphics[width=0.9\linewidth]{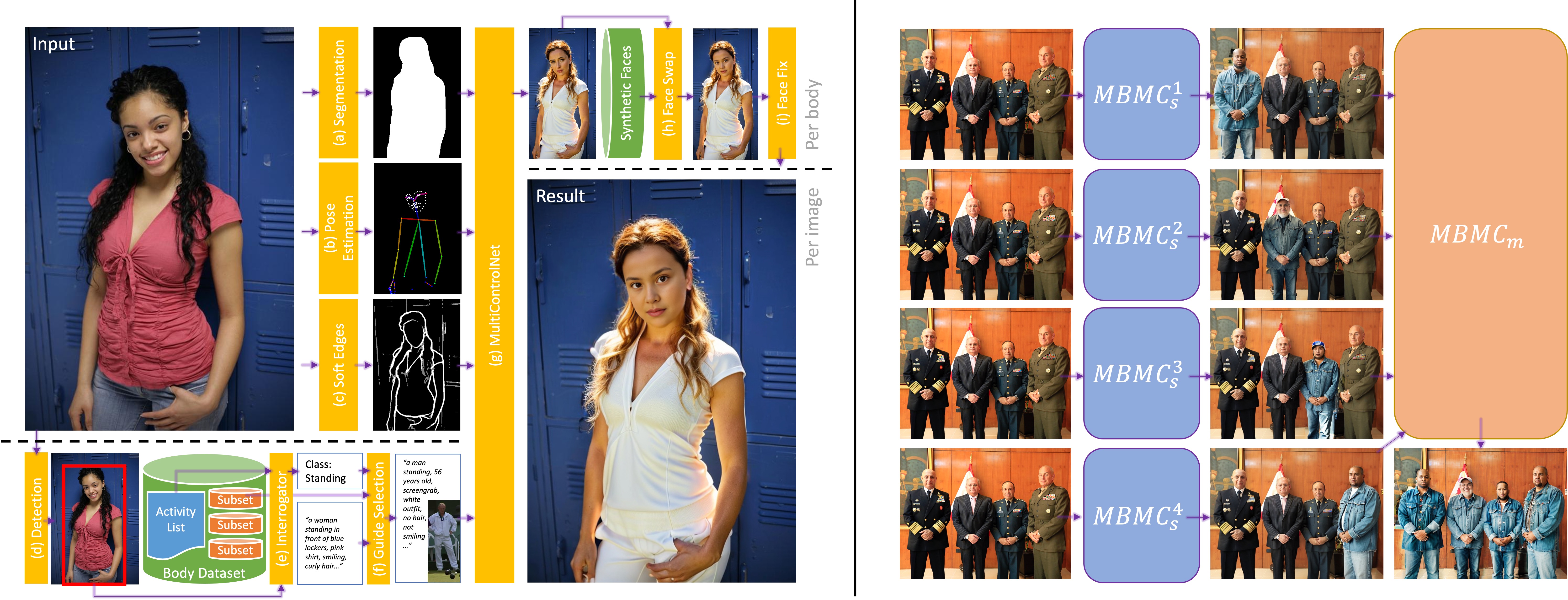}

   \caption{\textbf{Mask-based Removal Pipeline.} (Left) Embedding search, preparation of multiControlNet input, and face enhancements. (Right) Running the system independent of the body order for a multi-body image.}
   \label{fig:replace}
\end{figure*}

\section{Relevant Work}

Early approaches of privacy preservation may be reversible~\cite{reversible, reversible2} or result in over- or under-anonymization~\cite{anonfair}. Efforts have been emerging for content-preserving removal, defining removal as an inpainting task~\cite{regionfill,lama,gated,mat,instinpaint}. Beyond traditional image processing operations, various techniques such as style transfer~\cite{spade,adain,sean}, texture manipulation~\cite{Chen_2022_CVPR}, scene rearrangement~\cite{xu2023discoscene}, identity removal~\cite{mfmc}, and adversarial altercations~\cite{jia2022adv} are applied to replace sensitive features with plausible yet non-identifying content. VAEs and GANs are also explored via inpainting, content transfer, or part swap~\cite{pyramid,globallocal,context,structure}. With these, generative models solve the image quality issue, however, \textit{how} the anonymization is done is still an open research area. 

Most of the anonymization techniques focus on faces~\cite{ciagan,ganvideoanon,self,sun2018hybrid,Ren_2018_ECCV,sun2018natural}. 
Full body anonymization can address situations where contextual cues beyond the face play a significant role in identifying individuals. 
There are considerably less approaches focusing on body anonymization~\cite{shopping,survellience,surfaceguided, deepprivacy2}. Unfortunately, full-body anonymization almost always breaks the face structure and produces low quality results as seen in~\cite{surfaceguided,iknow, deepprivacy2}. 
\begin{table}[h]
\begin{tabular}{l|l|l|l|l|l}

Work       &\cite{ciagan}& \cite{lama}  & \cite{deepprivacy2}  & \cite{surfaceguided}& Ours  \\ \hline
Type      &inp &  inp     &   cloak    &   cloak         &  all         \\
Input    & img&    img  & img,corr   & img,corr &   im    \\
Scope & face &  both    & body   & body &  both     \\
Size & 128x128 &   any     &   288x160     & 384×256    &   any      \\ 
Guide & keypts &     mask     &   text     &  \cite{neverova2020continuous}   &    tasks     \\ 
Order & NA & seq & seq & seq & any\\
Train&   yes    &yes    &     yes      & yes    &   no      \\ 
\end{tabular}
 \caption{\textbf{Related Work.} We compare existing approaches based on their input and compute requirements.}
   \label{tab:ftw}
\end{table}

Some approaches guide anonymization with additional input~\cite{seginpaint,guidedsampling,nazeri2019edgeconnect}); define interesting loss functions~\cite{mfmc}; aim to replicate predefined priors~\cite{ciagan}, and let humans interfere~\cite{deepprivacy2}. 
All mentioned generative models used for anonymization are (1) trained from scratch; (2) on low-quality and low-diversity data; (3) sometimes with additional input; and (4) solving multi-body problem sequentially, order-dependent. We list these shortcomings of the SOTA in Tab.~\ref{tab:ftw}. My Body My Choice is a plug-and-play system, \textbf{does not need any training or fine-tuning}, works on any image size, does not require any additional input, is applicable to both body and face images, and supports order-invariant multi-body anonymization.

\section{My Body My Choice}
MBMC merges four anonymization paradigms with photorealistic results for each. Users have five options:
\vspace{-\topsep}
\begin{enumerate}\setlength{\parskip}{-2pt}
\vspace{2pt}
    \item \textbf{``Delete me from this photo.''} (physical removal)
    \item \textbf{``Do not auto-tag me.''} (adversarial removal) 
    \item \textbf{``Mask my body.''} (mask-based removal)
    \item \textbf{``Mask my face.''} (identity removal)
    \item \textbf{``No Action.''} 
\end{enumerate}

\subsection{System Overview}
MBMC options are designed as separate modules invoking multiple diffusion models or GANs for inpainting, body replacing, face replacing, and face enhancement tasks. All models are frozen, \textbf{only used for inference}, and no training or fine-tuning is performed. Operations are performed per body ($MBMC_s$) or per image ($MBMC_m$) as in Fig.~\ref{fig:replace}. 

\subsection{Physical Removal}
This option aims to completely eliminate the body from the image, filling in the cropped area with the background. Given an image $I$ with $n$ humans ${p_1, \dots, p_n}$ in it, physical removal employs YOLOv8~\cite{Jocher_YOLO_by_Ultralytics_2023} to segment each $p_i$ in $I$. We use inpainting model of ControlNetv1.1~\cite{ControlNetv1-1} where the control condition of the ControlNet is our segmentation mask with a fine-tuned Realistic Vision~\cite{RealisticVision} model of Stable Diffusion 1.5~\cite{rombach2021highresolution} backbone to synthesize masked regions with realistic attributes, patterns, or objects. 

\subsection{Adversarial Removal}
This option aims to eliminate or reduce the person detection accuracy as much as possible, which is called `vanishing gradient attack' in adversarial defense literature. We employ YOLOv3~\cite{redmon2018yolov3} with MobileNetv1~\cite{howard2017mobilenets} and DarkNet~\cite{redmon2018yolov3} backends as common detectors, operating on ``person'' class. We design an objectness gradient attack using TOG~\cite{TOG}, where the perturbation added to the image causes the gradients to vanish, creating no detection result. 
TOG targets real-time object detection systems, appropriate for scaling MBMC  
on a social media platform.   
Note that, it is possible to (1) create universal adversarial patches for all classes and (2) attack multiple detectors at the same time. We leave these extensions as future work.

\subsection{Mask-based Removal}
This option aims to replace the body with a quantifiably dissimilar body, utilizing the manifold of diffusion models, as illustrated in Fig.~\ref{fig:replace}. Given an image $I$ with $n$ humans ${p_1, \dots, p_n}$ in it, mask-based removal first employs YOLOv8~\cite{Jocher_YOLO_by_Ultralytics_2023} to segment each $p_i$ in $I$. The segmentation masks $M(p_i)$ are then dilated (Fig~\ref{fig:replace}a). Second, OpenPose~\cite{openpose} extracts pose skeleton and facial landmarks as $P(p_i)$ (Fig~\ref{fig:replace}b). Third, PiDiNet~\cite{su2021pixel} detects soft edges as $E(p_i)$ (Fig~\ref{fig:replace}c). Fourth, YOLOv8~\cite{Jocher_YOLO_by_Ultralytics_2023} is also used to detect the person bounding box as $B(p_i)$ (Fig~\ref{fig:replace}d).

The main novelty of MBMC is rooted in guiding the body synthesis. First, we create a body manifold as our embedding space with diverse bodies in terms of poses, activities, appearances, and demographics, thus, we select a subset of MPII Humans Dataset~\cite{mpii}. Every image $j_k$ comes with activity labels $A(j_k)$, so we store the same number of images ${j_1, \dots, j_m}$ from each activity class $t$. Then, we use CLIP Interrogator~\cite{inter} to compute the CLIP embeddings~\cite{clip} $C(j_k)$ of all images in this dataset (Fig~\ref{fig:replace}e). This manifold creation process is done only once and offline.

At run time, this time $B(p_i)$ is fed to the same interrogator, which returns the closest activity class $A(B(p_i))$ of the person and the CLIP embedding $C(B(p_i))$ for the person bounding box. The guide selection (Fig~\ref{fig:replace}f) searches for the $j_k$ satisfying $\forall j_k :  A(B(p_i))=A(j_k)$ and $\max_{j_k}\cos(C(B(p_i)),C(j_k)))$
where $\cos$ is cosine distance of embeddings $C(.)$, finding \textit{the furthest embedding within the same activity class}. 
We use a multi-ControlNet~\cite{ControlNetv1-1} architecture (Fig~\ref{fig:replace}g) based on Realistic Vision~\cite{RealisticVision} that combines multiple modalities. We input masked image $p_i-M(p_i)$, pose $P(p_i)$, soft edges $E(p_i)$ and CLIP embedding $C(j_{max})$ to this architecture to generate an image.

\subsection{Identity Removal}
Given an image $I$ with $n$ humans ${p_1, \dots, p_n}$ in it (or the multi-ControlNet output), identity removal applies ~\cite{mfmc} to find the face most dissimilar to $I$, within a randomness sphere. SimSwap~\cite{simswap} used in ~\cite{mfmc} causes under-anonymization based on less than expected drop in face recognition accuracy, so we use InSwapper~\cite{InsightFace} instead (Fig~\ref{fig:replace}h). Finally, 
we apply a face enhancer called GFPGAN~\cite{wang2021gfpgan} which increases face resolution and makes face details more prominent (Fig~\ref{fig:replace}i) by using the image priors that are encapsulated in a pretrained StyleGan2~\cite{Karras2019stylegan2} on face images. In order create faces with non-existing identities, we use a synthetic dataset~\cite{generatedphotos} for our embedding space. 

\subsection{Multi-body Anonymization}
Finally we summarize how everything works for multiple bodies (Fig~\ref{fig:replace}right). We explore two paths: Both of them run single-input MBMC on each person. The first one plugs back all results into the segmented areas, whereas the second one runs a multi-input MBMC with all results to let the diffusion model handle overlapping segmentations. As expected, the first approach creates order dependence on the persons, so we proceed with the second approach. 

\section{Results}
Evaluations are performed on an NVIDIA GeForce RTX 3070 with 8GB GPU, mask-based removal taking 30 sec and physical removal taking 20 sec. We use ~\cite{RealisticVision} as our diffusion checkpoint and DPM scheduler~\cite{lu2022dpmsolver} with 60 steps.
As our system does not require any training or fine-tuning, all listed datasets are used for evaluations. To evaluate MBMC in real world scenarios, so we gather an ambitious mix of diverse datasets as MPII Human Pose~\cite{mpii}, More Inclusive Annotations for People (MIAP)~\cite{miap}, People in Social Context (PiSC)~\cite{socialcontext}, Human Interaction Images (HII)~\cite{party}, Market 1501~\cite{market}, LaMa Humans~\cite{lama}, and generated.photos~\cite{generatedphotos}.

\begin{figure}[h]
  \centering
  \includegraphics[width=1\linewidth]{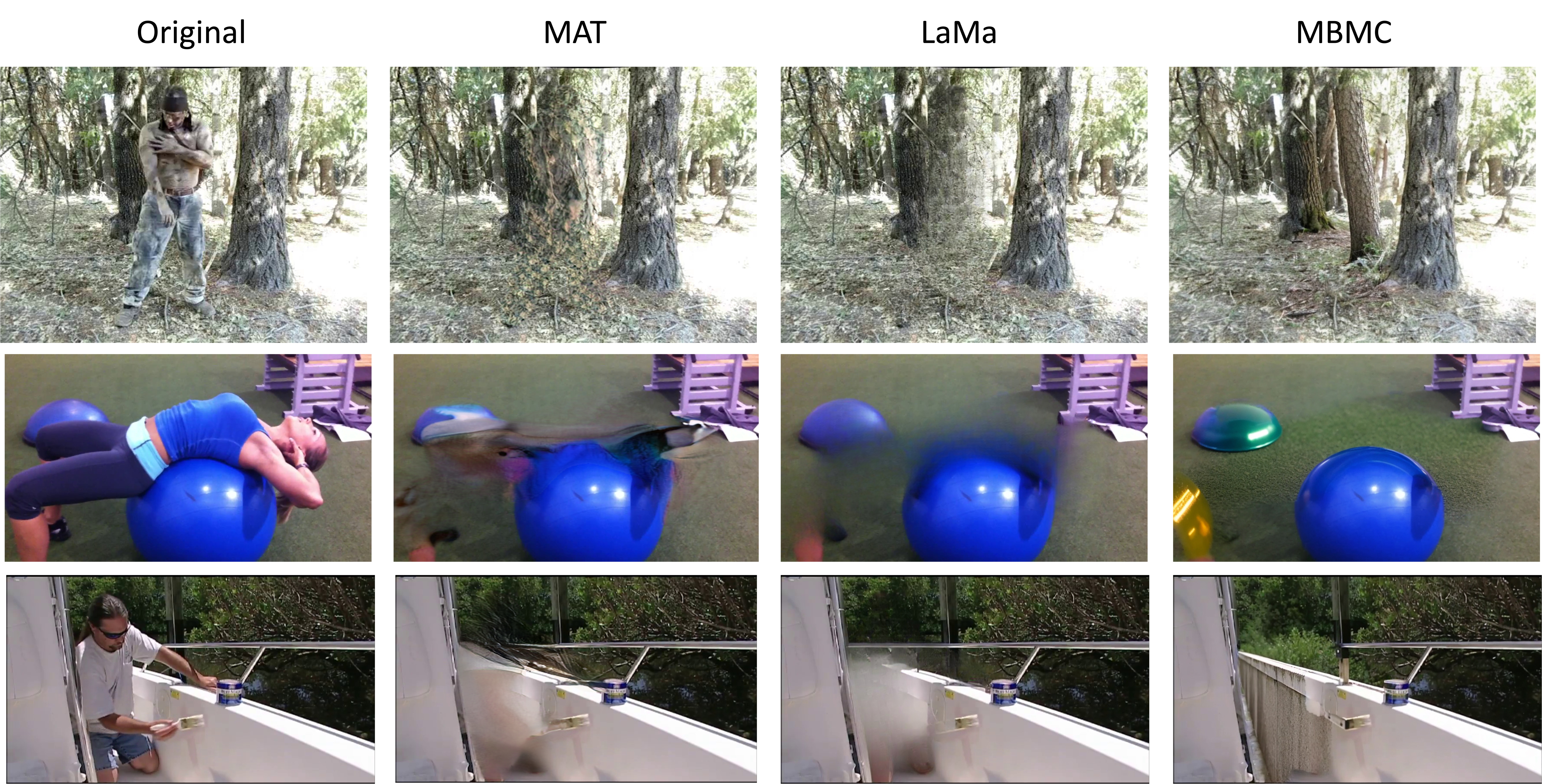}

   \caption{\textbf{Removal Comparison.} Compared to MAT~\cite{mat} and LaMa~\cite{lama}, MBMC creates less visual artifacts for patterned (top), extreme-pose (middle), and complex scene (bottom) cases.}
   \label{fig:removecomp}
\end{figure}
\subsection{Physical Removal}
We compare MBMC's anonymization by physical removal results to MAT~\cite{mat} and LaMa~\cite{lama} in Fig.~\ref{fig:removecomp} on MPII dataset. Instead of creating blurs in the forest or around the pilates ball, MBMC completes the images clearly with new content based on the diffusion model's understanding.

\subsection{Adversarial Removal}
We measure the detector accuracy before and after MBMC adversarial removal option on four datasets. In all of them, 97+\% accuracy is reduced to almost zero (Tab.\ref{tab:adv}). 
\begin{table}[h]
\centering
\begin{tabular}{l|l|l|l|l}

Dataset                                                   & LaMa & PiSC & MIAP  & MPII \\ \hline
Humans                       & 27         & 431           & 270   &    293     \\ \hline
Acc. before & 99.26      & 97.48         & 98.36 &     97.51   \\ \hline
Acc. after   & 0          & 0.07          & 0     &      0    \\ 
\end{tabular}
\caption{\textbf{Adversarial Removal.} Detector is fooled (0\% accuracy) on all four datasets after MBMC's adversarial removal.}
\label{tab:adv}
\end{table}

\subsection{Mask-based Removal}
\begin{figure}[h]
  \centering
  \includegraphics[width=1\linewidth]{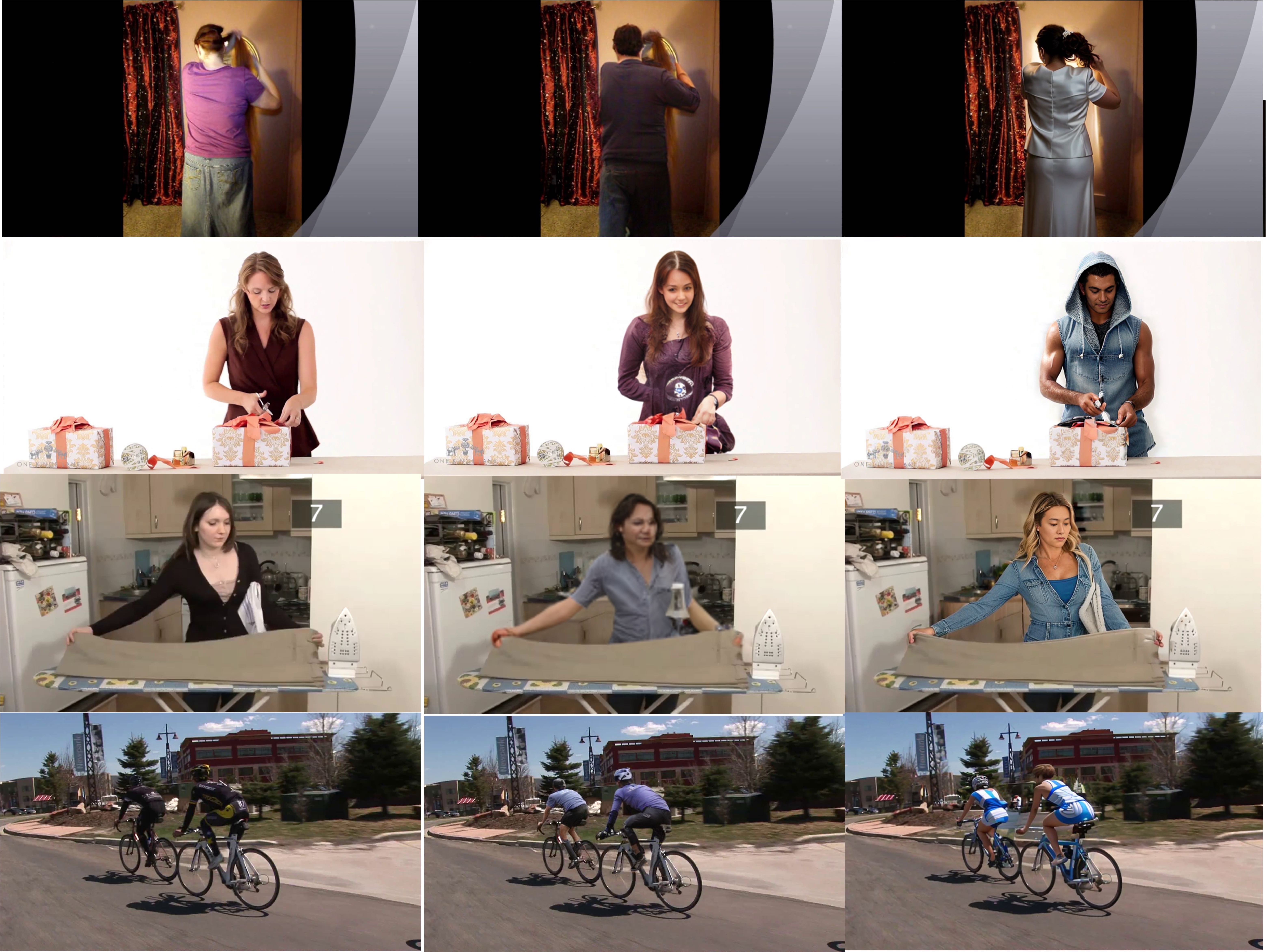}

   \caption{\textbf{Mask-based Removal Comparison.} Original,~\cite{deepprivacy2}, and ours are compared visually on MPII samples. Our results are more detailed, structured, contextual, and less similar to original.}
   \label{fig:dp}
\end{figure}

In Fig.~\ref{fig:dp}, we visually compare original, \cite{deepprivacy2}, and our results on MPII. Our results are of higher quality, contextually more coherent, and less similar to anonymized people in the images. Comparing FID scores, 99.92 (ours) vs. 73.00 (\cite{deepprivacy2}), we make more substantial changes. 

\begin{figure}[h]
  \centering
  \includegraphics[width=1\linewidth]{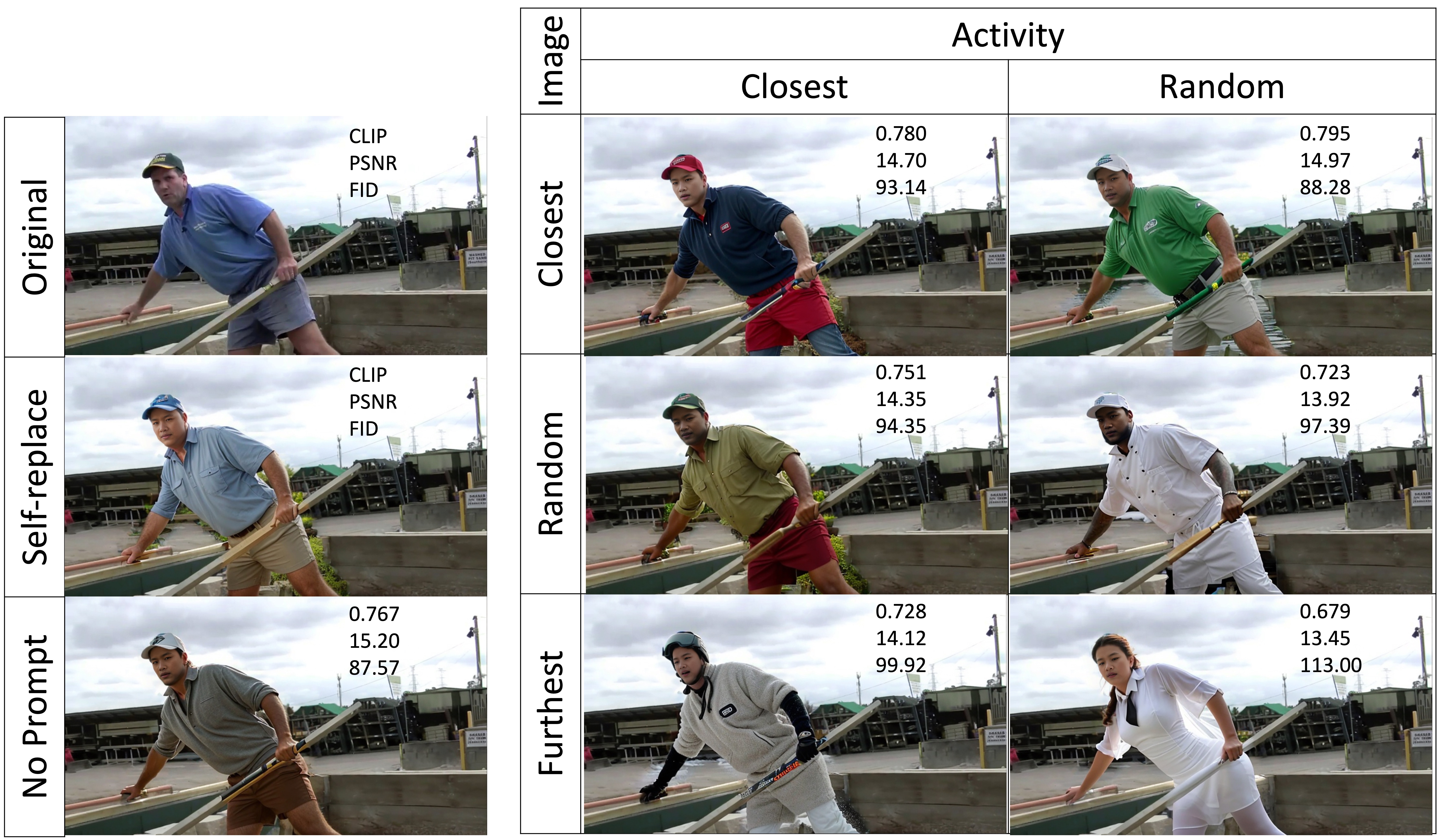}

   \caption{\textbf{Embedding Ablation.} MBMC currently uses the bottom mid, significantly different appearance with a similar pose.}
   \label{fig:embed}
\end{figure}

To motivate our embedding search for closest activity and furthest similarity image, we convey an ablation study with other possible embedding descriptions in Fig.~\ref{fig:embed}. CLIP, PSNR, and FID quantifications are reported to compare different embeddings. We also evaluate with no prompt and with the image's own embedding.

\begin{table}[h]
\begin{tabular}{l|l|l|l|l|l}
         & mAP & Rank1 & Rank5 & Rank10 & Rank20 \\ \hline
Original & 41.7  & 61.3   & 83.9   & 93.5    & 96.8    \\
MBMC     & 19.1  & 25.8   & 45.2   & 64.5    & 87.1   
\end{tabular}
\caption{\textbf{Re-identification after Mask-Based Removal.} Recognition accuracy significantly drops after our anonymization.}
\label{tab:reid}
\end{table}

\begin{figure}[h]
  \centering
  \includegraphics[width=.8\linewidth]{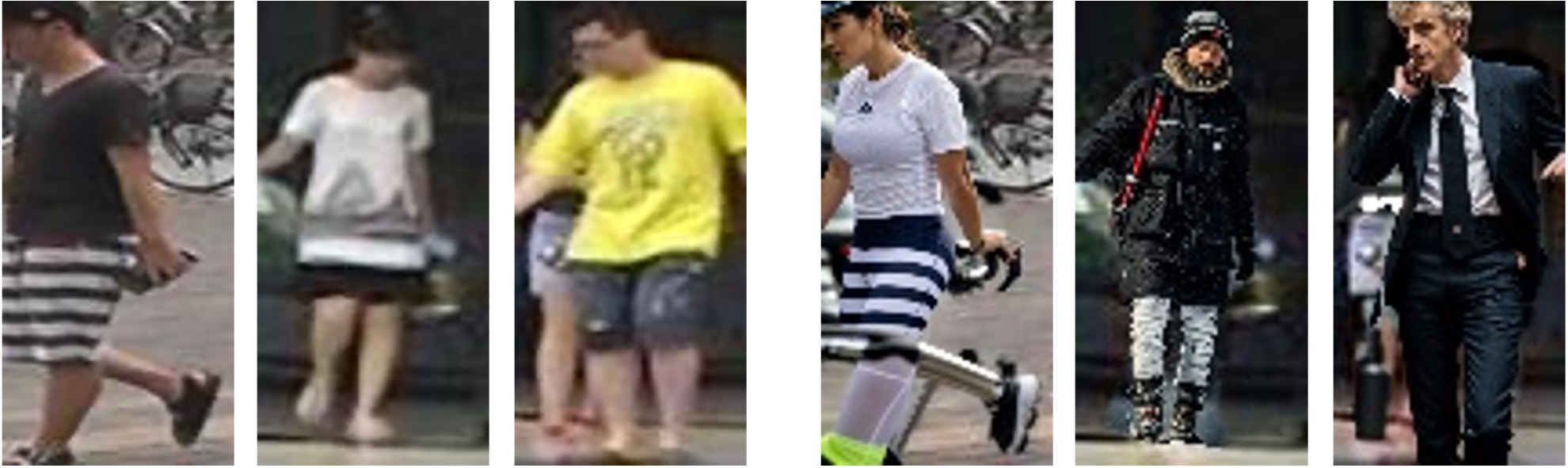}
   \caption{\textbf{Surveillance Anonymization.} MBMC works on any domain, size, resolution images, decreasing reidentification.}
   \label{fig:reid}
\end{figure}

We validate MBMC's mask-based removal by a re-identification experiment on Market 1501. In Tab.~\ref{tab:reid}, original precisions (first row) are significantly reduced (\%40+ reduction on Rank-1, second row) after MBMC is applied to surveillance images. Example images (left) and anonymized versions (right) are illustrated in Fig.~\ref{fig:reid}.

\subsection{Identity Removal}
We evaluate our face identity removal based on several datasets to asses its generalizability (Supp. A). After our modifications, ~\cite{mfmc} works much prevalently on hard cases. 

\section{Conclusion}

We present ``My Body My Choice'' to grant back the control over bodies to their owners. Our work extends the utility of diffusion models in anonymization by leveraging their capabilities with human guidance, with more results in Supp. B. As the call for ethical data handling and privacy preservation grows louder, our work marks a pivotal step towards responsible use of biometric data. 

{\small
\bibliographystyle{ieee_fullname}
\bibliography{egbib}
}

\end{document}


\title{My Body My Choice: Human-Centric Full-Body Anonymization\\ Supplemental Material}
\appendix
\author{Umur Aybars \c{C}ift\c{c}i\\
Binghamton University\\
{\tt\small uciftci@binghamton.edu}
\and
Ali Kemal Tanr{\i}verdi\\
Binghamton University\\
{\tt\small atanriv1@binghamton.edu}
\and
\.{I}lke Demir\\
Intel Labs\\
{\tt\small ilke.demir@intel.com}
}
\twocolumn[{%
\renewcommand\twocolumn[1][]{#1}%
\maketitle
\vspace{-0.5in}
\begin{center}
  \includegraphics[width=0.9\linewidth]{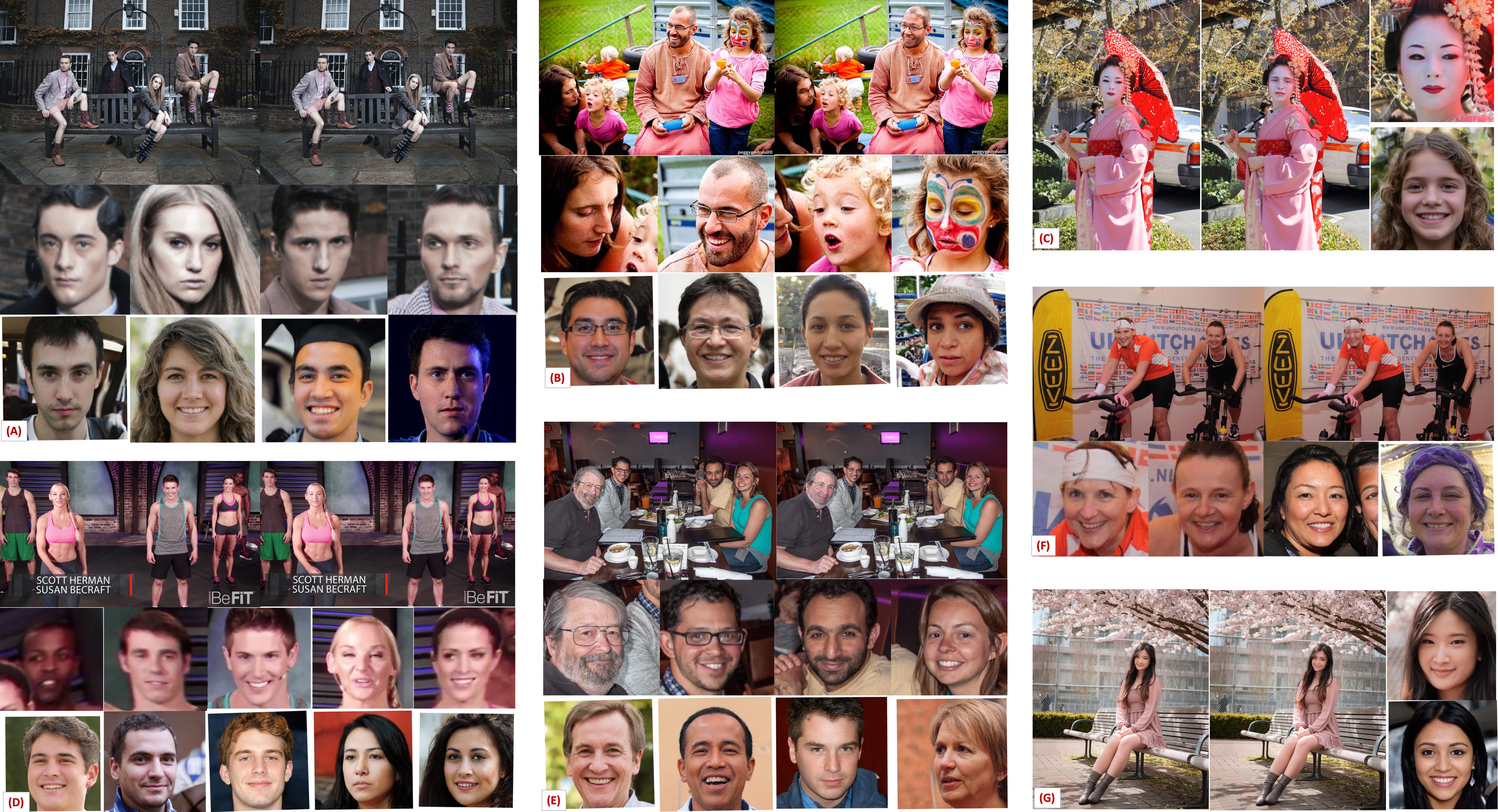}\\
   \textbf{Figure 1. }{\textbf{Identity Removal.}~\cite{mfmc} extension tested on four different datasets, with input output couples for images and faces.}
   \label{fig:mfmc}
\end{center}%
\stepcounter{figure}
}]
\maketitle

\section{Identity Removal}
Fig.1~\ref{fig:mfmc} demonstrates single and multi body identity removal samples from MPII (D), PiSC (B, E), MIAP (C, F) and Lama Humans (A, G) datasets, with original (left) anonymized (right) images and supporting source synthetic faces (small squares). We conclude that after our modifications, ~\cite{mfmc} works much prevalently on hard cases.

\begin{figure*}[hb]
  \centering
  \includegraphics[width=1\linewidth]{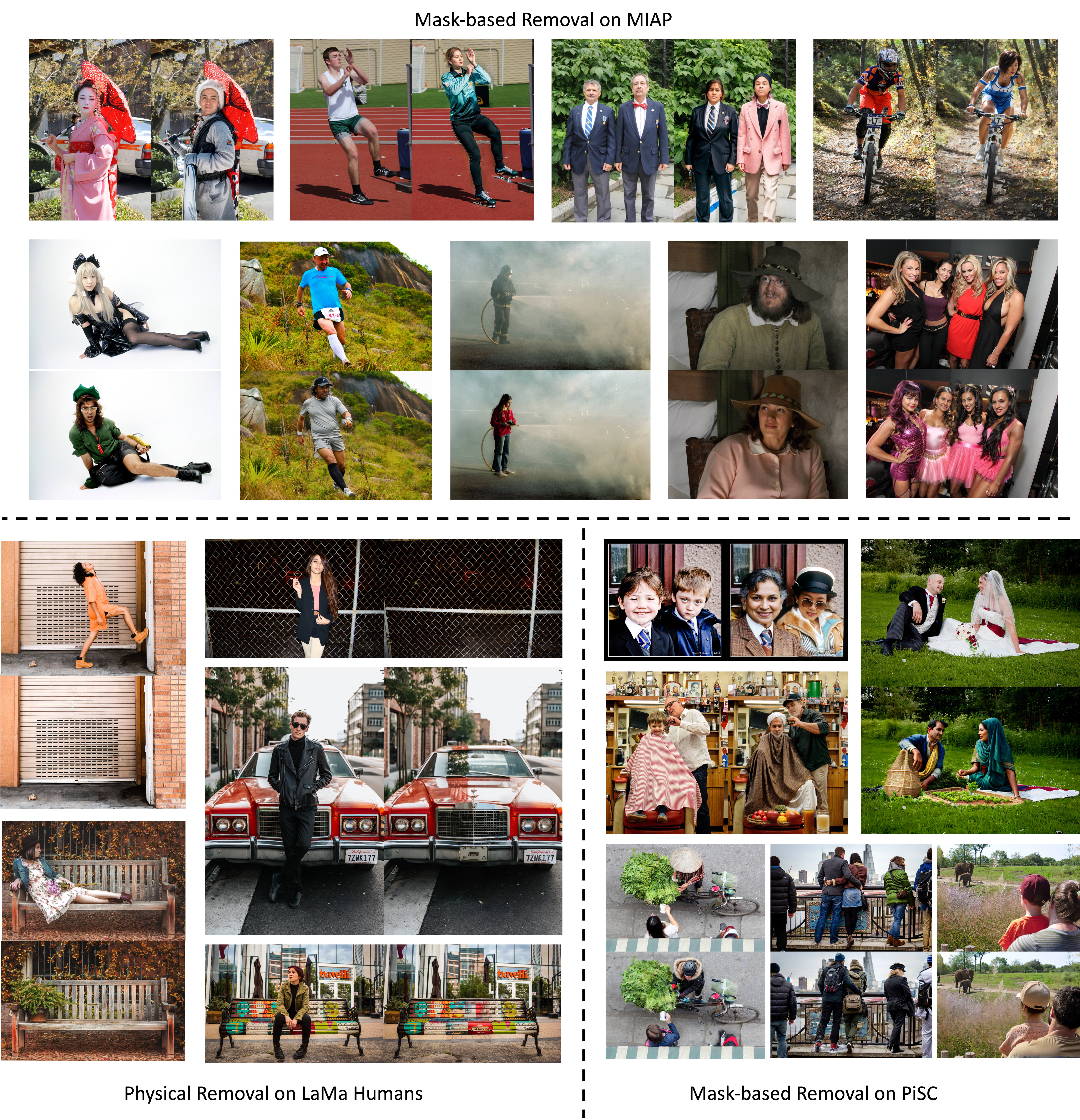} 

   \caption{\textbf{More Results.} Interesting example pairs for mask-based removal on MIAP~\cite{miap} (top) and PiSC~\cite{socialcontext}  (right), and for physical removal on LaMa Humans~\cite{lama} (left). All inputs are on the left or top of the pair and outputs are on the right or bottom.}
   \label{fig:all}
\end{figure*}

\section{All Dataset Results}
In Fig.~\ref{fig:all}, we select interesting results from MIAP~\cite{miap} (Fig.~\ref{fig:all}-top) and PiSC~\cite{socialcontext}  (Fig.~\ref{fig:all}-right) for mask-based, and LaMa Humans~\cite{lama} (Fig.~\ref{fig:all}-left) for physical removal task. Overall, images look realistic and people look significantly different; even in the existence of multiple people, complex illumination, fog, reflections, complex interactions, patterns, oblique views, occlusions, face paint, etc..

{\small
\bibliographystyle{ieee_fullname}
\bibliography{egbib}
}